\def\year{2024}\relax
\documentclass[letterpaper]{article} 
\usepackage{aaai22}  
\usepackage{times}  
\usepackage{helvet}  
\usepackage{courier}  
\usepackage[hyphens]{url}  
\usepackage{graphicx} 
\usepackage{natbib}  
\usepackage{caption} 
\DeclareCaptionStyle{ruled}{labelfont=normalfont,labelsep=colon,strut=off} 
\usepackage{algorithm}
\usepackage{algorithmic}
\usepackage{booktabs}
\usepackage{resizegather}
\usepackage{times}
\usepackage{epsfig}
\usepackage{graphicx}
\usepackage{amsmath}
\usepackage{amssymb}

\usepackage{newfloat}
\usepackage{listings}
\floatstyle{ruled}
\newfloat{listing}{tb}{lst}{}
\floatname{listing}{Listing}
\title{Decoupling the Curve Modeling and Pavement Regression for Lane Detection}
\author {
    Wencheng Han, Jianbing Shen
}
\affiliations{
    SKL-IOTSC, CIS, University of Macau \\
    {\tt\small \{wencheng256, shenjiangbingcg\}@gmail.com,}
}

\usepackage{bibentry}

\begin{document}

\maketitle

\begin{abstract}
The curve-based lane representation is a popular approach in many lane detection methods, as it allows for the representation of lanes as a whole object and maximizes the use of holistic information about the lanes. However, the curves produced by these methods may not fit well with irregular lines, which can lead to gaps in performance compared to indirect representations such as segmentation-based or point-based methods.
We have observed that these lanes are not intended to be irregular, but they appear zigzagged in the perspective view due to being drawn on uneven pavement. In this paper, we propose a new approach to the lane detection task by decomposing it into two parts: curve modeling and ground height regression. Specifically, we use a parameterized curve to represent lanes in the BEV space to reflect the original distribution of lanes. For the second part, since ground heights are determined by natural factors such as road conditions and are less holistic, we regress the ground heights of key points separately from the curve modeling.
Additionally, we have unified the 2D and 3D lane detection tasks by designing a new framework and a series of losses to guide the optimization of models with or without 3D lane labels. Our experiments on 2D lane detection benchmarks (TuSimple and CULane), as well as the recently proposed 3D lane detection datasets (ONCE-3Dlane and OpenLane), have shown significant improvements. We will make our well-documented source code publicly available.
\end{abstract}

\section{Introduction}
\label{Sec:intro}
Lane detection is a crucial task for autonomous driving, as it involves detecting traffic lanes in images to help with decision-making~\cite{bar2014recent,tang2021review,yenikaya2013keeping}. This field has gained significant attention recently~\cite{gebele2022carlane,zhang2022lane,xu2022rclane}. A long-standing question in this area is how to accurately represent lanes. Previous studies on lane detection~\cite{deng2022simultaneous,han2022laneformer,hou2019learning,neven2018towards} can be broadly classified into three categories: segmentation-based, point-detection-based, and curve-based methods. Segmentation-based methods~\cite{pan2018spatial,zheng2021resa} treat lane detection as a pixel-level classification task and identify lanes using heuristics in the post-processing stage. Point-based methods~\cite{zheng2022clrnet,liu2021condlanenet} detect lanes by locating a series of points and then interpolating them to form the lane. Both of these methods typically achieve state-of-the-art performance in this area, but they represent lanes in indirect ways, ignoring the lanes' inherent characteristics as a whole.

\begin{figure}[t]
  \centering
  \mbox{}\hfill
  \includegraphics[width = 0.99 \linewidth]{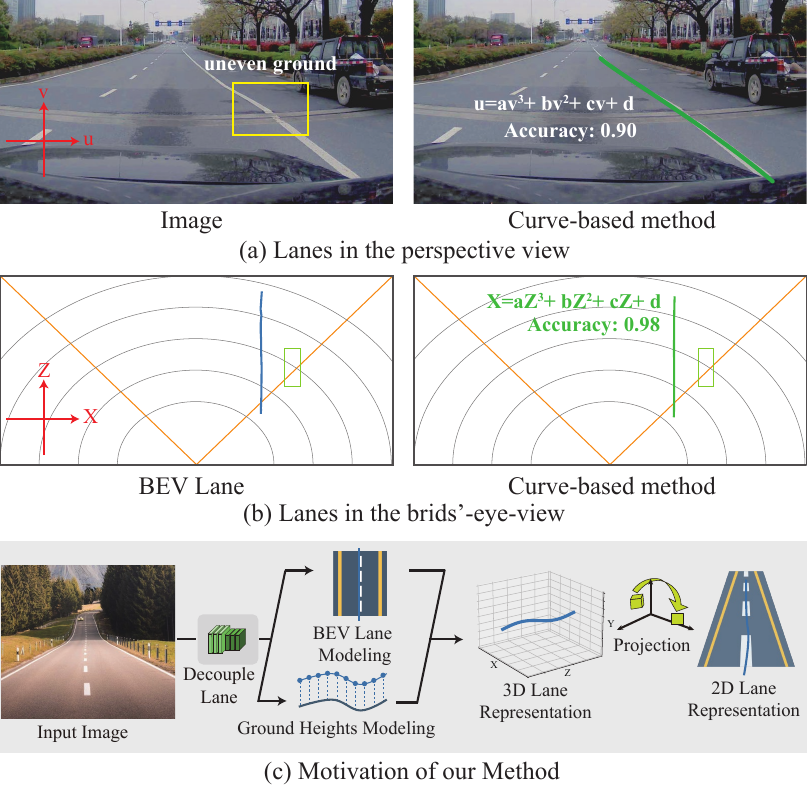}
\caption{\textbf{Illustration of a challenging case for curve-based lane representation and our solution.} (a) Uneven ground causes the lanes in the perspective view to fluctuate, making it difficult to fit the curves. (b) In the BEV space, the lanes retain their original status and can be easily fitted using parameterized curves. (c) Our solution is to separate the curve modeling and pavement regression and fit them independently.}
  \label{Fig:motivation}
\end{figure} 

Unlike the first two types of methods, curve-based methods take a holistic approach to representing lanes. Certain methods, such as those proposed by Van \textit{et al.}~\cite{van2019end} and Tabelini \textit{et al.}~\cite{tabelini2021polylanenet}, model lanes using polynomial curves, whereas Feng \textit{et al.}~\cite{feng2022rethinking} have proposed a method that utilizes B\'ezier curves to present lanes. These methods make full use of the geometric properties of lanes and present them elegantly, but they still perform less effectively than contemporary segmentation and point-based methods. Some researchers~\cite{feng2022rethinking} attribute this to the curve-based representations lacking enough degrees of freedom to accurately model lane shape. However, we believe that increasing the degrees of freedom even further may not be the optimal solution.

On one hand, freedom degrees and holistic representation are two opposing properties. If we increase the parameter numbers of the polynomial curves or the control points of the Bézier curves, the predictions may become more prone to overfitting local lane patterns. On the other hand, lanes are naturally designed to be simple and smooth in order to guide vehicle navigation. Thus, they should not be represented with too many complex fluctuations. After analyzing some challenging real-world cases of curve-based methods in 3D space, we discovered that most of the difficult lanes with irregular shapes in front-view images are actually due to the roads themselves, not the lanes. When viewed from above, many of these lanes exhibit smooth shapes that can easily be modeled using parameterized curves, as shown in Fig.~\ref{Fig:motivation}. For example, the lane in the yellow box appears irregular in the perspective view, making it difficult to fit accurately with a third-order polynomial. However, when viewed from above, the lane appears straight and can be easily modeled. Based on this observation, we conclude that uneven roads are the main cause of irregular lane shapes in the perspective view, rather than the lanes themselves.

To better represent the lanes, we propose a new framework called DecoupleLane. This framework represents the lanes in 3D space and models both the lanes and their corresponding ground heights. Afterwards, the lanes and ground heights are mapped back to the image space based on their perspective relationships.
Specifically, we focus on the original shape of lanes in the BEV space and use a polynomial to model the relationship between the $X$ and $Z$ coordinates. Unlike lane lines, the heights of roads are influenced by many natural factors and are not as holistic as the lanes. Therefore, following point-based methods, we represent the heights of the lane by a series of independent key points. Finally, the discrete values are interpolated and combined with $X$ and $Z$ values to form the 3D lanes.
Given the intrinsic matrix of the camera, we can convert the 3D lane coordinates into 2D ones, which can be used to represent lanes in the perspective view. For datasets that lack 3D lane labels, we train the model using the 2D loss between the converted 2D lanes and the 2D ground truth, as well as a regularization loss. This regularization constrains the ground heights to be as flat as possible, which helps the ground height head to focus on the irregular fluctuations of the road.
Although the predicted positions in the 3D space are not real coordinates, they can still encode the relative relationships between the lanes.
In conclusion, this paper's contributions can be summarized in four parts:

\begin{itemize}
    \item We thoroughly examined the limitations of curve-based representation and developed DecoupleLane to effectively utilize geometric information for comprehensive lane detection. This approach accurately accounts for fluctuations caused by uneven ground.
    \item We propose the Decouple Head for lane detection. It predicts the bird’s-eye view (BEV) representation and corresponding ground heights of the lanes. This approach disentangles the use of holistic information and complex environment information.
    \item We standardize the representation of lanes in both 2D and 3D spaces by treating the 2D lanes as a perspective-view projection of the 3D lane. We also propose a unified pipeline of loss to guide model optimization.
    \item Our method outperforms state-of-the-art approaches on two representative 2D lane detection datasets: TuSimple and CULane. It also shows strong performance on recent 3D lane detection datasets, including ONCE-3DLane and OpenLane.
\end{itemize}

\section{Related Works}

\subsection{2D lane representation.}
The 2D lane detection task involves locating lanes in a 2D space. Based on different lane representations, most 2D lane detection models can be sorted into three categories: segmentation-based methods, point-based methods, and curve-based methods.

\noindent\textbf{Segmentation-based methods}
This approach treats lane detection as a task of classifying each pixel and assigning it a specific lane class. One pioneering study by Pan~{\textit{et al.}}~\cite{pan2018spatial} utilized slice-by-slice convolution within feature maps and allowed for message-passing between pixels across rows and columns in a layer. Hou \textit{et al.}~\cite{hou2019learning} introduced knowledge distillation and proposed self-attention distillation to enhance the performance of lightweight models. Their approach achieved comparable performance to state-of-the-art models while using significantly fewer parameters. Zheng {\textit{et al.}}~\cite{zheng2021resa} implemented a recurrent feature-shift aggregator to enhance lane feature extraction and designed a bilateral up-sampling decoder to combine coarse and fine-grained features during the upsampling procedure. Ghafooria {\textit{et al.}}~\cite{ghafoorian2018gan} suggested a generative model to improve the realism and structure preservation of the prediction. Although segmentation-based approaches have achieved satisfactory performance, they often require substantial post-processing due to the misrepresentation of the data. Additionally, the representation disregards the overall context of the lane, which makes it challenging to utilize holistic information.

\noindent\textbf{Point-based methods}
Inspired by object detection methods, point-based models approach lane detection as a prediction problem of points, where lanes are represented by a series of points.
Li \textit{et al.}~\cite{li2019line} were the first to introduce a point-based representation in this area and used predefined anchors to provide prior information.
Tabelini {\textit{et al.}}~\cite{tabelini2021keep} employed an attention module to improve the model's performance on challenging scenes, such as occlusion and missing lane markers, by enhancing global information extraction.
Qin {\textit{et al.}}~\cite{qin2020ultra} treated the lane detection task as a row-based selection problem with global features and introduced a structural loss to explicitly model the structure of lanes.
Zhang {\textit{et al.}}~\cite{zheng2022clrnet} presented the cross-layer refinement network, which fully utilizes both high-level and low-level features in lane detection. They introduced the line IoU loss to regress the lane line globally, resulting in improved localization accuracy.
Wang {\textit{et al.}}~\cite{wang2022keypoint} directly regressed the keypoints by predicting the offsets to the starting point of the lane line. They proposed a Lane-aware Feature Aggregator to capture the local correlations between adjacent keypoints. This representation achieves state-of-the-art performance, although the indirect modeling of lanes still makes it less efficient to utilize holistic information.

\noindent\textbf{Curve-based methods}
Curve-based methods use curve functions to predict a series of parameters and represent lane lines.
Van \textit{et al.} \cite{van2019end} introduced this type of representation, which consists of two components: a deep network for weight map prediction and a differentiable least squares fitting module to produce the best-fitting curves.
Tabelini \textit{et al.} \cite{tabelini2021polylanenet} used deep polynomial regression to produce a polynomial representation for each lane marking.
Liu \textit{et al.} \cite{liu2021end} developed a transformer-based method that extracts abundant features and directly outputs parameters of a lane shape model.
Feng \textit{et al.} \cite{feng2022rethinking} proposed a parametric B\'ezier curve to represent the lane lines.
Curve-based methods are considered the most intuitive way to represent lanes because they can easily model the lane distribution with rectified curves.
However, current curve-based methods do not perform as well as contemporary segmentation and point-based methods, possibly because they cannot handle complex environmental changes such as uneven ground.
In this paper, we identify this problem and suggest that more research is needed to improve this method's performance.

\subsection{3D lane representation.}
3D lane detection is a new task in the lane detection field that requires accurate information about the lane in 3D space instead of just the image space. This introduces some new challenges in lane representation compared to 2D lane detection.
The pioneering work by Garnett {\textit{et al.}}~\cite{garnett20193d} introduced intra-network inverse-perspective mapping (IPM) and anchor-based lane representation to solve the 3D lane detection task. Guo~\textit{et al.}~\cite{guo2020gen} subsequently designed a geometry-guided lane anchor in the virtual BEV and decoupled the learning of image segmentation and 3D lane prediction. Yan~\textit{et al.}~\cite{yan2022once} proposed a new real-world 3D lane detection dataset and an end-to-end detector.
The proposed network SALAD is extrinsic-free and anchor-free, which regresses the 3D coordinates of lanes in image view without explicitly converting the feature map into the BEV. Bai {\textit{et al.}}~\cite{bai2022curveformer} presented a one-stage Transformer-based method that directly predicts 3D lane parameters.
Chen {\textit{et al.}}~\cite{chen2022persformer} used a unified 2D/3D anchor design to detect 2D/3D lanes simultaneously. They also released a large-scale real-world 3D lane dataset called OpenLane.
These works have tried different types of representation but all consider 3D lane representation as an integrated target, ignoring the complex environment and intuitive lane design.

\section{Method}
In this section, we will first introduce the structure of the proposed DecoupleLane in \S \textbf{Overview}. 
Then in \S \textbf{Decouple Head}, we will explain the core module of our method in detail, the Decouple Head, which disentangles the representation of lanes into holistic BEV lane modeling and discrete key point height regressions. 
In \S \textbf{Unified Lane Detection}, we unify the 3D and 2D lane representation and treat the 2D lane as a perspective projection of the corresponding 3D lane. 
%

\subsection{Overview}
\label{Sec:overview}
\begin{figure}[t]
  \centering
  \mbox{}\hfill
  \includegraphics[width = 0.99 \linewidth]{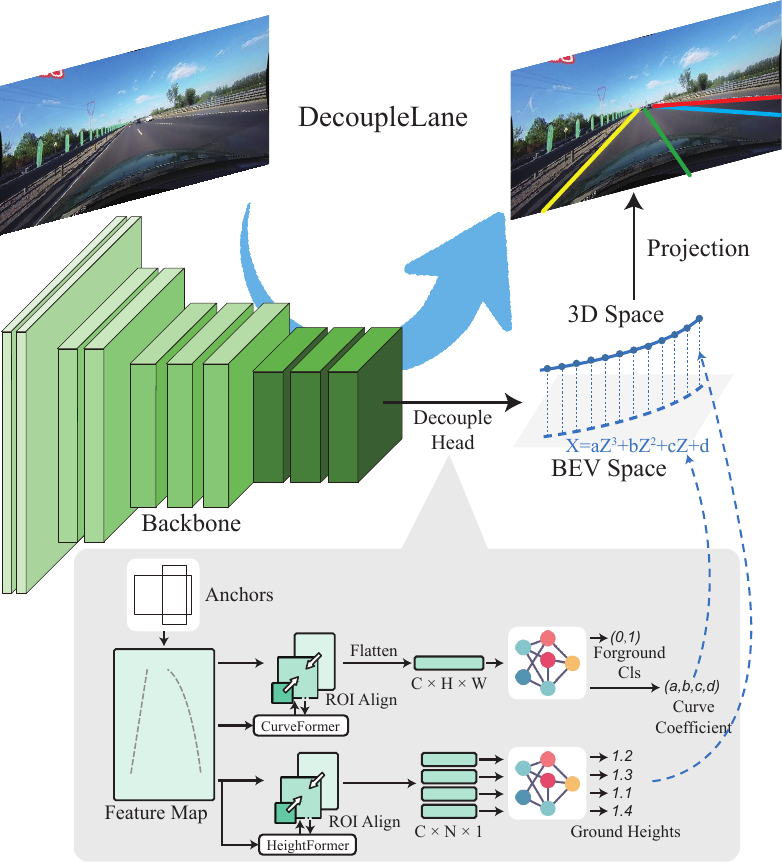}
  \hfill\mbox{}  \mbox{}\hfill
\caption{\textbf{An overview of the proposed DecoupleLane Framework.} There are mainly two parts to the proposed network, a backbone for feature extraction and the Decouple Head for decoding the features into lane representation.}
  \label{Fig:network}
\end{figure}

Fig.~\ref{Fig:network} shows an illustration of the proposed DecoubleLane. Different from previous curve-based methods~\cite{feng2022rethinking,zheng2021resa}, which usually pool the feature maps into one row and employ a head network to decode the curve in the corresponding column, our model employs an anchor-based architecture. 
Although column-based methods are efficient in computation, this architecture still has some limitations in terms of 3D lane representation. Firstly, column-wise pooling can not handle the side lanes well. 
As shown in Fig.~\ref{Fig:former} (a), the two lanes on the right side are pooled into the same column, making it hard for the model to discriminate them. 
Secondly, as we discussed in \S\textbf{Introduction}, ground height is a critical factor that influences the representation of the lane. 
The pooled row features can not encode abundant spatial information like the ground heights and thus can not fulfill our requirement. 
As shown in Fig.~\ref{Fig:network}, there are mainly two parts in the model, a backbone network for the feature extraction and a Decouble Head for decoding the feature into the corresponding 3D lanes. 
In this paper, we employ a DLA-34~\cite{yu2018deep} network as our backbone, which can efficiently extract high-level semantic information while keeping a high resolution of the feature map. 
The advantages will help our model to capture holistic information like the curve shapes of the lane and discriminate the detailed features like the ground undulation. 

Then, we define several anchors by the clustering algorithm for the lanes and employ ROI-Align modules to pool the features into fixed shapes. 
Unlike the object detection task, the distribution of lanes is more reasonable, and a small number (less than 50) of anchors can recall most of the lanes. 
Also, different from point-based methods, we do not need the anchors to provide holistic priors because our curve representation can achieve this implicitly. Therefore, our anchor design is a simple but effective way to extract possible lane features $x$:
\begin{equation}
x = ROIAlign(f(\gamma, I)),
\end{equation}
where $I$ is the input front view image and $\gamma$ is the weight of the backbone network.

The features gathered are then sent to the Decouple Head for lane decoding. This process models the lane curve in the BEV space and predicts the corresponding ground heights. It is worth noting that we do not need to manually convert the image into the top view or extract features like some previous works~\cite{garnett20193d,guo2020gen}. This is because the IPM conversion is based on the flattened ground hypothesis, which can introduce additional noise to the reasonable lane representation in the BEV space. In our work, all the features are directly extracted from the input 2D images. Finally, curves and heights are combined to produce the 3D lanes. Our DecoupleLane can also generate 2D lane representations. Instead of using two separate heads for 2D and 3D representation, our model projects the 3D lanes into the perspective view to produce 2D lanes. More information about this process is available in \S\textbf{Unified Lane Detection}.

\subsection{Decouple Head}
\label{Sec:decouple}
As discussed in \S\textbf{Introduction}, there are two main factors that affect the shape of the lane in the perspective view. The first factor is the design of the lane itself, which is typically created by experts and therefore has a reasonable and elegant shape. This is why the lane is considered to be holistic. The second factor is the ground on which the lane is situated. The height of the ground is influenced by various natural factors such as topographical changes, which causes irregular fluctuations. These two factors together make it difficult to fit the lane's projection in the perspective view.

\begin{figure*}[t]
  \centering
  \includegraphics[width = 0.6 \linewidth]{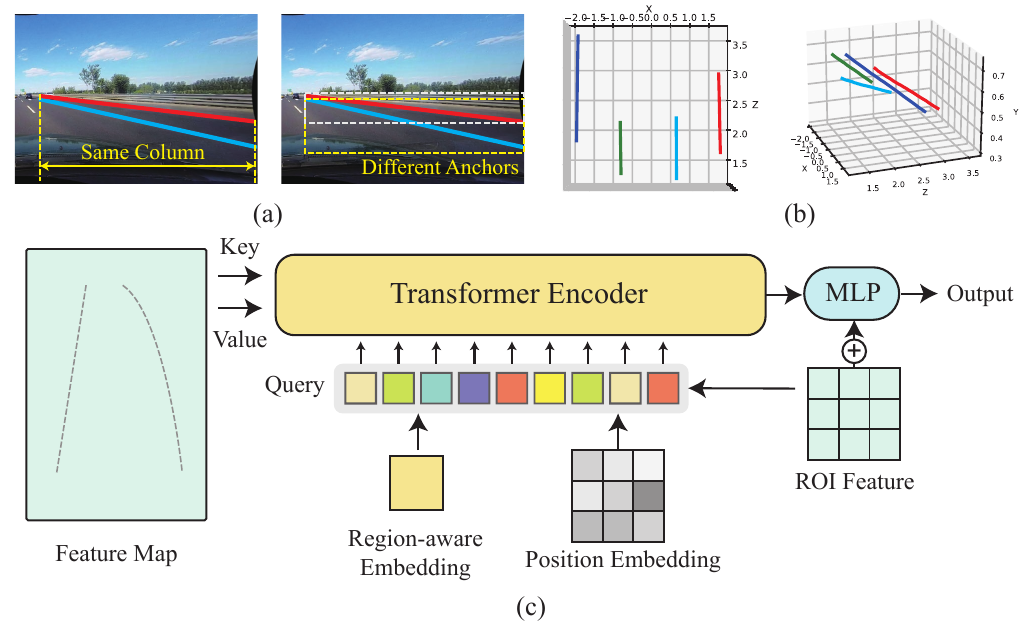}
\caption{\textbf{Illustration of the lanes and the structure of the CurveFormer and HeightFormer.} (a) Illustration of the aligned problem of column-pooling operation. (b) Illustration of the 3D representation of DecoupleLane trained with only 2D labels. (c) The structure of the CurveFormer and HeightFormer.}
  \label{Fig:former}
  \vspace{-2mm}
\end{figure*}

To fully utilize the geometry information and effectively handle the two factors, we propose the Decouple Head. This model separately models the lanes in the BEV space and the ground heights. In the BEV space, we focus on the relationships between the $\mathcal{X}$ and $\mathcal{Z}$ coordinates while ignoring the $\mathcal{Y}$ (height) value changes in a Left-hand Cartesian System. As mentioned earlier, lanes in the BEV have holistic and elegant shapes. We prefer a simple representation without large degrees of freedom in this procedure.
Therefore, we employ a third-order polynomial to model the $\mathcal{X}, \mathcal{Z}$ relationship:
\begin{equation}
\mathcal{X} = a\mathcal{Z}^3 + b\mathcal{Z}^2 + c\mathcal{Z} + d,
\end{equation}
where $a$, $b$, $c$, $d$ are the predicted coefficients of the curve. 
To calculate the coefficients with a comprehensive perspective of the lanes, we start by utilizing an ROI-Align module on the feature maps using the predetermined anchors. This process transforms the features into a fixed size. Next, we create a new CurveFormer to collect global information. Our CurveFormer is based on a transformer structure, as illustrated in Fig.~\ref{Fig:former} (c). It receives the ROI feature as queries and the global feature map as keys and values.

To distinguish the relevant information, we have equipped the ROI Features with two kinds of embeddings. Firstly, there is the position embedding which encodes the relative position information of the ROI Features. Secondly, there is the learnable region-aware embedding for each anchor. These region-aware embeddings are optimized during the end-to-end training process, and they help the queries focus on the overall information of the curve.
Finally, we flatten the output features of the CurveFormer and send them into a fully-connected layer for predicting the coefficients:

\begin{equation}
[a, b, c, d]^\intercal = f(\omega, flatten(CurveFormer(x))),
\end{equation}
where $\omega$ is the weight of the fully-connected layer. 

In addition to the BEV lanes, accurately modeling the ground heights $\mathcal{Y}$ is another crucial aspect. Changes in ground height are affected by natural factors like ground conditions, making it challenging to fit them into a parameterized model. To tackle this issue, we adopted a point-based approach. First, we ROI-Align the input feature into a column shape $\bar{x} \in (1, n)$, where $n$ is a hyper-parameter that determines the number of key points. In our experiments, we set it to 72. Then, we aggregate local information from the corresponding key-point positions using a HightFormer. The proposed HightFormer has a structure similar to CurveFormer but is trained independently, enabling it to encode different priors during training and focus on disparate regions from the CurveFormer.
Finally, we apply a fully-connected layer on every pixel of the feature map and convert it into the height of points:
\begin{equation}
\mathcal{Y}_i = f(\eta, HeightFormer(\Bar{x})_i).
\end{equation}

We uniformly select $n$ key points on the lane along the Z-axis. Each value of $\mathcal{Y}_i$ represents the height of the corresponding key point, as illustrated in Fig~\ref{Fig:network}. Next, we interpolate the heights to create a continuous sequence and then merge the BEV lane representation with the heights to generate the 3D representation.

\subsection{Unified Lane Detection}
\label{Sec:unified}
Previous works have often treated 2D and 3D lane detection as separate tasks, with two types of lanes represented independently~\cite{chen2022persformer}. However, it is important to note that 2D and 3D representations ultimately refer to the same lanes in the real world, and treating them separately may lead to limitations. For example, in 2D representation, perspective distortion can cause undesirable deformation of lanes. When two parallel lanes intersect at the vanishing point, their relationship can be difficult to comprehend. Furthermore, due to the different representations, large-scale datasets and training methods for 2D lane detection may not be directly applicable to improving the performance of 3D lane detection, and vice versa, making the use of data less efficient.

{To bridge the two tasks, our method uniformly represents lanes in the 3D space, and then project 3D lane representations into the perspective view to construct the 2D lane predictions.} 
Given a lane point $P=(\mathcal{X}, \mathcal{Y}, \mathcal{Z})$ in the 3D space and the intrinsic matrix $K$ of the camera, the corresponding 2D point $p=(u, v)$ can be calculated by:
\begin{equation}
\mathcal{Z}\left[ \begin{array}{c}
        u \\
        v \\
        1
        \end{array} 
        \right ] = \left[ \begin{array}{ccc}
        f_x & 0 & o_x\\
        0 & f_y & o_y\\
        0 & 0 & 1
        \end{array} 
        \right ] \left[ \begin{array}{c}
        \mathcal{X} \\
        \mathcal{Y} \\
        \mathcal{Z}
        \end{array}
        \right ],
\end{equation}
where $f_x$ and $f_y$ are the pixel focal lengths and $o_x, o_y$ are the offsets of the principal point.
Then we will guide the optimization of the model in both 3D and 2D spaces.

\noindent \textbf{Unified Loss} To create the loss, we start by aligning the predicted lanes with the ground truth lanes. Our approach involves aligning them in 2D space, rather than 3D space, because feature maps are generated from 2D images. This makes it easier for the model to extract features accurately based on the 2D locations.
Specifically, we employ the Hungarian algorithm to assign predictions to the ground truth 2D lanes, and the costs are formulated as:
\begin{equation}
    C = \mathcal{L}_1(u_p, u_g) + \mathcal{L}_1(v_{sp}, v_{sg}) + \mathcal{L}_1(v_{ep}, v_{eg}),
\end{equation}
where $\mathcal{L}_1(u_p, u_g)$ are the average horizon distance between the predicted lanes and the ground truth lanes and $\mathcal{L}_1(v_{sp}, v_{sg}), \mathcal{L}_1(v_{ep}, v_{eg})$ are the vertical distance of the corresponding start and end points. 
Next, we assign positive labels to successfully matched predictions and negative labels to the rest. We then use a classification loss, $L_{cls}$, to guide the discrimination between the two groups.

For the 3D lanes, we design the loss in two parts. Specifically, to guide the optimization of the holistic lane curve, we employ a Lane IoU loss~\cite{zheng2022clrnet} in the BEV space:
\begin{equation}
L_{bev}=\frac{2e + \min \left(\mathcal{X}_i^p, \mathcal{X}_i^g\right)-\max \left(\mathcal{X}_i^p, \mathcal{X}_i^g\right)}{2e + \max \left(\mathcal{X}_i^p, \mathcal{X}_i^g\right)-\min \left(\mathcal{X}_i^p, \mathcal{X}_i^g\right)},
\end{equation}
where $e$ is a hyper-parameter that controls the radius of the lanes and $\mathcal{X}_i$ is the corresponding $\mathcal{X}$ values of the uniformly sampled points. 
We use an $L_1$ loss to compare the predicted $\mathcal{Y}$ values with the ground truth, which is referred to as the height loss $L_h$. Additionally, we employ an endpoint $L_Z$ loss to estimate the error between the $\mathcal{Z}$ values of the start and end points:
\begin{equation}
L_{3D} = L_{bev} + L_h + L_{Z}
\end{equation}

\begin{figure*}[t]
  \centering
  \includegraphics[width = 0.8 \linewidth]{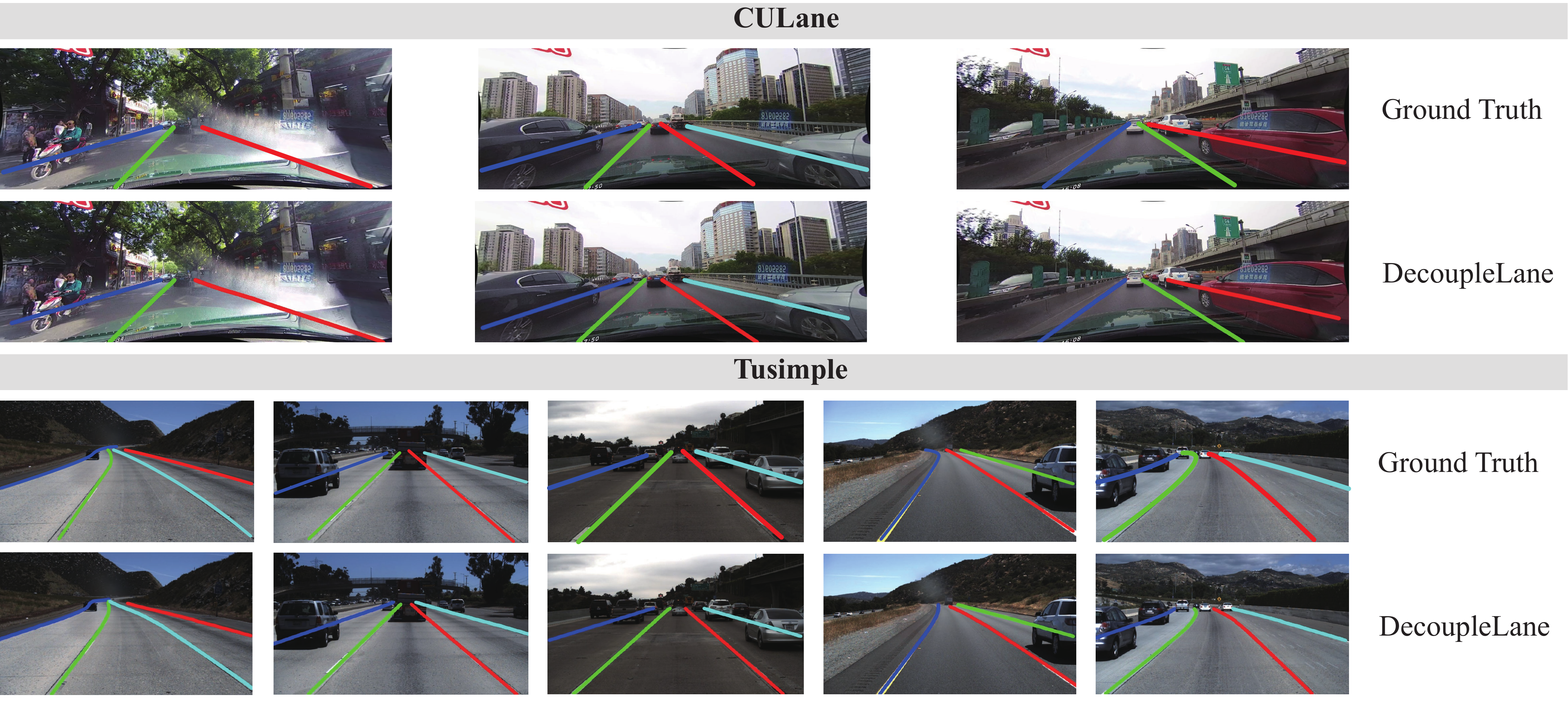}
\caption{\textbf{Visualization of the prediction of  DecoupleLane.} The visualization of the predictions of DecoupleLane and the corresponding ground truth on CULane and TuSimple are shown. The color of the lane is employed to discriminate different instances.}
  \label{Fig:vis_2d}
\end{figure*}

For the 2D lanes, we employ a Lane IoU loss $L_{per}$ in the perspective space and an endpoint loss $Lv$ for estimating the error between $v$ values of the start and end points in the 2D space:
\begin{equation}
L_{2D} = L_{per} + L_v
\end{equation}
Totally our loss is formulated as:
\begin{equation}
    L = \left\{  \begin{array}{ll}
        	L_{cls} + \alpha L_{3D} + \beta L_{2D} ,&  \text{if 3D labels available} \\
        L_{cls} + \beta L_{2D} + |\sigma_h|,&  \text{otherwise}
    \end{array} 
\right., 
\end{equation}
where $\alpha$ and $\beta$ are the balance ratio and in our experiments. 
When training with data that doesn't have 3D labels, such as the 2D lane detection datasets mentioned earlier, we remove the 3D losses since there are no corresponding 3D labels available. However, doing so can cause the relationship between the BEV representation and the height regression to become unstable. To address this issue, we add an additional regulation loss $|\sigma_h|$ to minimize the variance of the ground heights while optimizing the parameters.
In this setting, the 3D representation can be regarded as a implicit representation representation for 2D lane detection. 
{Although the model can not produce accurate 3D lane locations without 3D guidance, 
based on the perspective theory, the model can learn reasonable 3D relationships under only 2D guidance as shown in Fig.~\ref{Fig:former} (b).}

\section{Experiments}

\begin{table*}

    \begin{center}

        \resizebox{\textwidth}{!}{%

            \begin{tabular}{@{}lrrrrrrrrrrrrr@{}}

                \toprule

                \multicolumn{1}{c}{\textbf{Method}} & \multicolumn{1}{c}{\textbf{mF1}} &
                \multicolumn{1}{c}{\textbf{F1@50}} &
                \multicolumn{1}{c}{\textbf{F1@75}} &
                \multicolumn{1}{c}{\textbf{FPS}} &
                \multicolumn{1}{c}{\textbf{Normal}} & \multicolumn{1}{c}{\textbf{Crowded}} & \multicolumn{1}{c}{\textbf{Dazzle}} & \multicolumn{1}{c}{\textbf{Shadow}} & \multicolumn{1}{c}{\textbf{No line}} & \multicolumn{1}{c}{\textbf{Arrow}} & \multicolumn{1}{c}{\textbf{Curve}} & \multicolumn{1}{c}{\textbf{Cross}} & \multicolumn{1}{c}{\textbf{Night}} \\ 
                \hline
                \multicolumn{14}{c}{{Segmentation-based Method}} \\
                \hline
                SCNN(VGG16)~\cite{pan2018spatial} & 38.84 & 71.60 & 39.84 & 8  & 90.60 & 69.70 & 58.50 & 66.90 & 43.40 & 84.10 & 64.40 & 1990 & 66.10 \\
                
                RESA(ResNet34)~\cite{zheng2021resa}  & - & 74.50 & 45.50 & 41 & 91.90 & 72.40 & 66.50 & 72.00 & 46.30 & 88.10 & 68.60 & 1896 & 69.80  \\

                RESA(ResNet50)~\cite{zheng2021resa} & 47.86 & 75.30 & 35.70 & 43 & 92.10 & 73.10 & 69.20 & 72.80 & 47.70 & 88.30 & 70.30 & 1503 & 69.90  \\

                UFLD(ResNet18)~\cite{qin2020ultra}  & 38.94 & 68.40 & 40.01 & \textbf{282} & 87.70 & 66.00 & 58.40 & 62.80 & 40.20 & 81.00 & 57.90 & 1743 & 62.10 \\

                UFLD(ResNet34)~\cite{qin2020ultra}  & - & 72.30 & - & 170 & 90.70 & 70.20 & 59.50 & 69.30 & 44.40 & 85.70 & 69.50 & 2037 & 66.70 \\

                \hline
                \multicolumn{14}{c}{{Point-based Method}} \\
                \hline
                FastDraw(ResNet50)~\cite{philion2019fastdraw} &  - & - & 90.30 & - & 85.90 & 63.60 & 57.00 & 69.90 & 40.60 & 79.40 & 65.20 & 7013 & 57.80  \\
                
                PINet~\cite{ko2021key} & 46.81 & 74.40 & 51.33 & 25 & 90.30 & 72.30 & 66.30 & 68.40 & 49.80 & 83.70 & 65.20 & 1427 & 67.70 \\
                
                LaneATT~\cite{tabelini2021keep} & 49.57 & 76.68 & 54.34 & 129  & 92.14 & 75.03 & 66.47 & 78.15 & 49.39 & 88.38 & 67.72 & 1330 & 70.72 \\

                LaneAF~\cite{abualsaud2021laneaf} & 50.42 & 77.41 & 56.79 & 20 & 91.80 & 75.61 & 71.78 & 79.12 & 51.38 & 86.88 & 72.70 & 1360 & 73.03 \\

                SGNet~\cite{su2021structure}  & - & 77.27 & 92.00 & - & 92.07 & 75.41 & 67.75 & 74.31 & 50.90 & 87.97 & 69.65 & 1373 & 72.69 \\
                
                FOLOLane~\cite{qu2021focus}  & - & 78.80 & 40.00 & - & 92.70 & 77.80 & 75.20 & 79.30 & 52.10 & 89.00 & 69.40 & 1569 & 74.50 \\

                CondLane(ResNet34)\cite{liu2021condlanenet} & 53.11 & 78.74 & 59.39 & 128 & 93.38 & 77.14 & 71.17 & 79.93 & 51.85 & 89.89 & 73.88 & 1387 & 73.92 \\

                CondLane(ResNet101)\cite{liu2021condlanenet}  & 54.83 & 79.48 & 61.23 & 47 & 93.47 & 77.44 & 70.93 & 80.91 & 54.13 & 90.16 & 75.21 & 1201 & 74.80 \\

                CLRNet(ResNet34)\cite{zheng2022clrnet}  & 55.14 & 79.73 & 62.11 & 103 & 93.49 & 78.06 & 74.57 & 79.92 & 54.01 & 90.59 & 72.77 & 1216 & 75.02 \\

                CLRNet(DLA34)\cite{zheng2022clrnet}  & 55.64 & 80.47 & 62.78 & 94 & 93.73 & 79.59 & 75.30 & \textbf{82.51} & 54.58 & \textbf{90.62} & 74.13 & 1155 & 75.37 \\
                \hline
                \multicolumn{14}{c}{{Curve-based Method}} \\
                \hline
                LSTR\cite{liu2021end}  & - & 68.72 & - & 47 & 86.78 & 67.34 & 56.63 & 59.82 & 40.10 & 78.66 & 56.64 & 1166 & 59.92 \\
                
                B\'ezierLaneNet(ResNet-18)\cite{feng2022rethinking}  & - & 73.67 & - & 213 & 90.22 & 71.55 & 62.49 & 70.91 & 45.30 & 84.09 & 58.98 & 996 & 68.70 \\

                B\'ezierLaneNet(ResNet-34)\cite{feng2022rethinking} & - & 75.57 & - & 150 & 91.59 & 73.20 & 69.20 & 76.74 & 48.05 & 87.16 & 62.45 & \textbf{888} & 69.90 \\

                \hline
                \textbf{DecoupleLane(ours)} & \textbf{56.32} & \textbf{80.82} & \textbf{63.25} & 110 & \textbf{93.85} & \textbf{80.32} & \textbf{76.31} & {82.34} & \textbf{55.40} & {90.48} & \textbf{75.12} &1035 & \textbf{75.40} \\
                
                \bottomrule
            \end{tabular}

        } %

    \end{center}
    \caption{
    Comparison with other state-of-the-art lane detectors on CULane benchmark. For a fair comparison, the fps metric of DecoupleLane in this table is reported on a single GTX 1080ti GPU.
    }
    \vspace{-2mm}
    \label{tab:culane_main}

\end{table*}

\subsection{Implementation Details}
Most of our experiments are conducted on a platform with an Intel Platinum 8260 CPU, 30GB RAM, and a single Tesla V100 GPU. When testing the inference speed of the DecoupleLane, we run our model on a single GTX 1080ti GPU for a fair comparison with other methods.  During our training and inference, all the pixels above the skylines in the images are removed to save unnecessary computation and the inputs are resized into $800 \times 320$. For data augmentation, we employ random horizontal flips, translation, and scaling. Besides processing the images and 2D labels, we also adjust the corresponding intrinsic matrix to keep the relationship between the 2D and 3D lanes. We employ an AdamW optimizer with a learning rate of 1e-3 and a cosine decay learning rate strategy with a power of 0.9. For CULanes, TuSimple, ONCE-3DLane, and OpenLane, we train our model on their training set with 15,70,10,10 epochs respectively.

\subsection{Dataset}
To show the efficiency of the proposed method, we conduct our experiments on two representative 2D lane detection benchmarks, TuSimple~\cite{Tusimple} and CULane~\cite{pan2018spatial}. Also, we evaluate our method on two recently proposed 3D lane detection datasets ONCE-3DLane~\cite{yan2022once} and OpenLane~\cite{chen2022persformer}. 

\noindent\textbf{TuSimple} is one of the most popular benchmarks in this area, which contains 3,268 images in the training set, 358 for validation, and 2,782 for testing. All of the images are captured in the highway scenes and are with the resolution of $1280 \times 720$.

\noindent\textbf{CULane} contains 9 different challenging factors \textit{e.g.} crowded, night, cross \textit{etc.} to evaluate the robustness of the model. Also, it is a large-scale dataset with a total of 88,880 frames for the training set, 9675 for the validation set, and 34680 for the test set.

\noindent\textbf{ONCE-3DLane} is a recently proposed 3D lane detection benchmark, which is built on the large-scale autonomous driving dataset ONCE~\cite{mao2021one} which contains more than one million scenes. ONCE-3DLane manually labeled the 2D lances and automatically generated the corresponding 3D lane labels based on the LiDAR point clouds. This dataset contains 5,000 scenes for the training set and 3,000 for the validation set and 8,000 scenes for the testing set. 

\noindent\textbf{OpenLane} is a large-scale realistic 3D lane detection benchmark built on the well-known Waymo Open dataset~\cite{sun2020scalability}. OpenLane contains more than 200,000 frames and over 880,000 carefully annotated lanes, and each of the frames could contain as many as 24 different lanes.

\noindent\textbf{Metrics} F1-measure is the most commonly used metric in this area and is employed as the main metric in CULane~\cite{pan2018spatial}, ONCE-3DLane~\cite{yan2022once} and OpenLane~\cite{chen2022persformer}. Firstly, all the lanes are assumed as 30 pixels wide and the IoU values between the predicted lanes and the ground truth are calculated. Based on this, predictions with IoU values large than a given threshold (0.5 as default) are thought to be positive, and the F1 metric is defined as:
\begin{equation}
F_1=\frac{2 \times \text { Precision } \times \text { Recall }}{\text { Precision }+\text { Recall }} \text {, }\nonumber
\end{equation}
where Precision$ =\frac{T P}{T P+F P}$ and Recall$ =\frac{T P}{T P+F N}$. Following~\cite{zheng2022clrnet}, to evaluate the overall performance of the models, we employ the mF1 metric like COCO~\cite{lin2014microsoft} detection dataset:
\begin{equation}
\mathrm{mF} 1=(\mathrm{F} 1 @ 50+\mathrm{F} 1 @ 55+\cdots+\mathrm{F} 1 @ 95) / 10, \nonumber
\end{equation}
where $\mathrm{F} 1 @ p$ is the $F1$ values with $p\%$ IoU thresholds. For TuSimple~\cite{Tusimple} Accuracy is employed as the main metric:
\begin{equation}
\text { Accuracy }=\frac{C_{pred}}{N_{gt}}, \nonumber
\end{equation}
where $C_{pred}$ is the corrected number in a predicted lane and $N_{gt}$ is the number of lanes in the gt.

\begin{table}
    \begin{center}
    \resizebox{0.5\textwidth}{!}{%
            \begin{tabular}{@{}lcccc@{}}
                \toprule
                \textbf{Method}      & \textbf{F1 (\%)} & \textbf{Acc (\%)}  & \textbf{FP (\%)} & \textbf{FN (\%)} \\ \midrule
                \multicolumn{5}{c}{{Segmentation-based Method}} \\
                \hline
                SCNN~\cite{pan2018spatial}  & 95.97 & 96.53  & 6.17 & \textbf{1.80} \\
                
                RESA~\cite{zheng2021resa}  & 96.93 & 96.82 & 3.63 & 2.48  \\

                UFLD\cite{qin2020ultra}   & 88.02 & 95.86  & 18.91 & 3.75 \\
            
                \hline
                \multicolumn{5}{c}{{Point-based Method}} \\
                \hline

                LaneATT(ResNet34)\cite{tabelini2021keep}  & 96.77 & 95.63 & 3.53 & 2.92 \\
                
                LaneATT(ResNet122)\cite{tabelini2021keep}  & 96.06 & 96.10 & 5.64 & 2.17 \\

                CondLaneNet(ResNet34)\cite{liu2021condlanenet} & 96.98 & 95.37  & 2.20 & 3.82 \\
                
                CondLaneNet(ResNet101)\cite{liu2021condlanenet}  & 97.24 & 96.54 & \textbf{2.01} & 3.50 \\

                CLRNet(ResNet34)\cite{zheng2022clrnet} & 97.82 & 96.87  & 2.27 & 2.08 \\
            
                \hline
                \multicolumn{5}{c}{{Curve-based Method}} \\
                \hline
                PolyLaneNet~\cite{tabelini2021polylanenet}   & 90.62 & 93.36 & 9.42 & 9.33 \\
                B\'ezierLan(ResNet-18)~\cite{feng2022rethinking} & - & 95.41  & 5.30 & 4.60 \\
                
                B\'ezierLan(ResNet-34)~\cite{feng2022rethinking}  & - & 96.54 & 5.10 & 3.90 \\
               
                \midrule
                DecoupleLane(ours)  & \textbf{97.93} & \textbf{97.01} & 2.03 & 3.31 \\
                \bottomrule
            \end{tabular}
    }

    \end{center}
    \caption{State-of-the-art results on TuSimple. Additionally, F1 was computed using the official source code.}
    \label{tab:tusimple_main}

\end{table}

\subsection{State-of-the-art Comparison}
\noindent\textbf{CULane}
Table~\ref{tab:culane_main} shows our comparison with the state-of-the-art method on CULane. We group the methods into three categories based on their representation types. Among all the methods, our DecoupleLane achieves the best performance and set a new state-of-the-art performance. Fig.~\ref{Fig:vis_2d} show some visualization comparison with other methods on CULane. Compared with other methods, DecoupleLane fits the lanes smoothly and easily handle some challenging scene like occlusion.

\noindent\textbf{TuSimple} 
Table~\ref{tab:tusimple_main} shows our comparison on the Tusimple benchmark. Compared with CULane, Tusimple has more curving lanes. Our method still performs better than all of the comparing methods and sets a new start-of-the-art performance.

\begin{figure*}[t]
  \centering
  \begin{center}
  \includegraphics[width = 0.7 \linewidth]{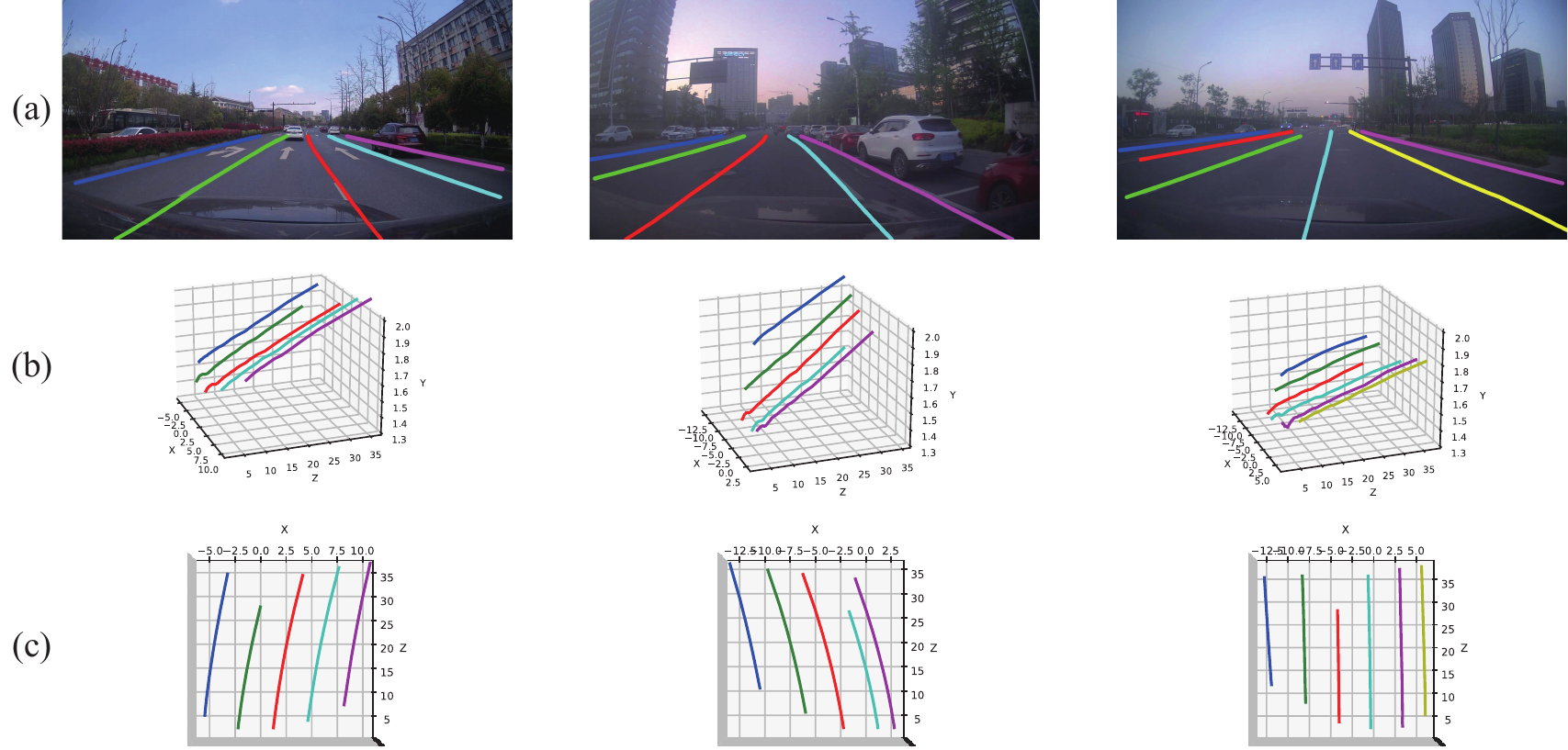}
  \end{center}
\caption{\textbf{Illustration of the predictions on the ONCE-3DLane benchmark.} (a) The projected lane lines in the perspective view. (b) Illustration of the 3D lanes in a Left-hand Cartesian System. (c) Illustration of the 3D lanes in the BEV space.}
  \label{Fig:vis_3d}
  \vspace{-3mm}
\end{figure*}

\noindent\textbf{ONCE-3DLane}
For quantitative comparison, as ONCE-3DLane does not lease its labels in the testing set or the online evaluation server, we report comparison on the validation set in Table~\ref{tab:once_3dlance}. As shown in this table, our DecoupleLane outperforms all of the other state-of-the-art 3D lane detection methods and achieves 75.07\% of the F1 score.
For the qualitative comparison, Fig.~\ref{Fig:vis_3d} shows the illustrations of the proposed DecoupleLane. Fig.~\ref{Fig:vis_3d} (a) shows the projected 2D lanes in the front image, which can ideally match the real lanes in the image, even though the ground is not flat. Fig.~\ref{Fig:vis_3d} (b) (c) shows the 3D lanes in a Left-hand Cartesian System and the BEV space. From these figures, we can find that although the lanes show fluctuating status in the perspective view and the 3D space because of the uneven ground, our model can discriminate their original relationships in the BEV space.

\noindent\textbf{OpenLane}
To show the generality of our method, we also employ OpenLane for comparison. Table~\ref{tab:openlane} shows that our method can work with this dataset well and outperform other newly proposed methods.

\begin{table}
    \begin{center}
    \resizebox{0.45\textwidth}{!}{%
            \begin{tabular}{@{}lrrrrrr@{}}
                \toprule
                \textbf{Method}     & \textbf{F1 (\%)} & \textbf{Precision (\%)}  & \textbf{Recall (\%)} & \textbf{CD error (m)} \\ \midrule
                3D-LaneNet~\cite{garnett20193d} & 44.73 & 61.46 & 35.16 &  0.127 \\               Gen-LaneNet~\cite{guo2020gen} & 45.59 & 63.95 &  35.42 &  0.121 \\
                PersFormer~\cite{chen2022persformer} & 74.33 & 80.30 &  69.18 &   0.074 \\
                Anchor3DLane~\cite{huang2023anchor3dlane} & 74.44 & 80.50 &  69.23 &   0.064 \\
                DecoupleLane & \textbf{75.07} & \textbf{81.19}  & \textbf{69.26} & \textbf{0.062} \\

                \bottomrule
            \end{tabular}
    }
    \end{center}
    \caption{Performance of the proposed DecoupleLane on ONCE-3DLane validation set.}

    \label{tab:once_3dlance}

\end{table}

\begin{table}
    \begin{center}
    \resizebox{0.4\textwidth}{!}{%
            \begin{tabular}{@{}lccccc@{}}
                \toprule
                \textbf{Method} & \textbf{All} & \textbf{Up Down} & \textbf{Curve}  & \textbf{Night} & \textbf{Intersection} \\ \midrule
                
                3D-LaneNet~\cite{garnett20193d} & 44.1 & 40.8 & 46.5  & 41.5 & 32.1 \\
                
                Gen-LaneNet~\cite{guo2020gen} & 32.3 & 25.4 & 33.5 & 18.7 & 21.4  \\
                
                PersFormer~\cite{chen2022persformer} & 50.5 & 42.4 & 55.6 & 46.6 & 40.0 \\
                CurveFormer~\cite{bai2022curveformer} & 50.5 & \textbf{45.2} & 56.6 & \textbf{49.1} & 42.9 \\
                DecoupleLane &\textbf{51.2} & 43.5 & \textbf{57.3} & 48.9 & \textbf{43.5} \\

                \bottomrule
            \end{tabular}
    }

    \end{center}
    \caption{State-of-the-art results on OpenLane.}

    \label{tab:openlane}
    \vspace{-3mm}
\end{table}

\begin{table}[t]
	\centering
	\addtolength{\tabcolsep}{0pt}
	\resizebox{0.45\textwidth}{!}{%
	\begin{tabular}{*{11}{c}}
		\toprule
		\textbf{Unify} & \textbf{Decouple} & \textbf{Former} & \textbf{mF1}   & \textbf{F1@50}  & \textbf{F1@75} \\
		\midrule
		                      &                          &                  &  51.90 & 78.37 & 58.32    \\
		\checkmark            &                          &                  &  52.80 & 78.27 &  59.50    \\
		\checkmark           & \checkmark               &                  &  54.74 & 78.91  & 61.77     \\
		\checkmark            & \checkmark               & \checkmark       & \textbf{56.32} & \textbf{80.82} & \textbf{63.25}   \\
		\bottomrule
	\end{tabular}
	}
	\caption{Ablation studies of each component in our method. Results are reported on CULane. }
	\label{tab:ablation}
	
\end{table}

\begin{table}
    \begin{center}
    \resizebox{0.45\textwidth}{!}{%
            \begin{tabular}{@{}lrrrrr@{}}
                \toprule
                \textbf{Method}  & \textbf{mF1 } & \textbf{F1@50}  & \textbf{F1@75} \\ \midrule
                
                second-order polynomial & 51.25 & 76.32  & 57.36  \\
                
                third-order polynomial & \textbf{56.32} & \textbf{80.82} & \textbf{63.25}  \\
                
                fourth-order polynomial  & 56.15 & 80.32 & 63.14 \\

                B\'ezier curves  & 55.26 & 80.14 & 62.60  \\
                
                \bottomrule
            \end{tabular}
    }

    \end{center}
    \caption{Ablation study of different curve representation methods. Results are reported on CULane.}

    \label{tab:curve_method}
    \vspace{-3mm}
\end{table}

\subsection{Ablation Study}
In this section, we study the influence of the proposed three modifications \textit{e.g.}, the unified lane detection structure~(Unify), the DecoupleHead~(Decouple), and the Curve/HeightFormer module~(Former). For the baseline model, we employ a point-based 2D lane detection head on our backbone model and employ the same anchors to extract feature maps. Table~\ref{tab:ablation} show the efficiency of the corresponding modules. According to this table, all three components provide a positive effect on the performance, and DecoupleHead contributes the most proportion of improvement. Besides the modules, we also compare different curve representations in Table~\ref{tab:curve_method}. We first compare polynomials with different orders, the third-order polynomial performs better than the second-order polynomial, but the fourth-order polynomial performs comparably with the third-order polynomial.  B\'ezier curves show no improvements with the polynomials in our Decouple Head. Therefore, to achieve the best balance between performance and efficiency, we choose the third-order polynomial as our curve representation.

\section{Conclusion}
In this paper, we analyze the challenges of curve-based lane detection methods and find that irregular lanes in the perspective view are caused by ground fluctuations. To address this issue, we propose DecoupleLane, which better fits lane shapes and fully utilizes the holistic representation of parameterized curves. The Decouple Head, a core module in our method, models the curve in the BEV space and regresses the ground heights separately. Additionally, DecoupleLane unifies the 2D and 3D lane detection tasks by employing a single 3D lane detection head and considering the 2D lanes as projections in the perspective space. This approach improves the model's understanding of the real distribution of lanes in the 2D lane detection task, and the large amount of 2D lane data improves the robustness of the model in the 3D lane detection task. We evaluate our method on two representative 2D lane detection benchmarks and two 3D lane detection datasets, and our method achieves state-of-the-art performance in all cases.

{\small
\bibliography{egbib}
}

\end{document}